\documentclass[conference]{IEEEtran}
\IEEEoverridecommandlockouts

\usepackage{cite}
\usepackage{amsmath,amssymb,amsfonts}
\usepackage{algorithmic}
\usepackage{graphicx}
\usepackage{textcomp}
\usepackage{xcolor}
\usepackage{booktabs}
\usepackage{multirow}
\usepackage{url}
\usepackage{tikz}
\usetikzlibrary{arrows.meta,positioning,fit,backgrounds}
\usepackage{subcaption}

\def\BibTeX{{\rm B\kern-.05em{\sc i\kern-.025em b}\kern-.08em
    T\kern-.1667em\lower.7ex\hbox{E}\kern-.125emX}}

\begin{document}

\title{A Dual-Branch Graph and Convolutional Network for Interpretable Movement EEG Classification}

\author{
\IEEEauthorblockN{Maryam Ostadsharif Memar}
\IEEEauthorblockA{
\textit{Department of Electrical and Computer Engineering} \\
\textit{Isfahan University of Technology} \\
Isfahan, Iran \\
m.ostadsharif@ec.iut.ac.ir
}
\and
\IEEEauthorblockN{Behzad Nazari}
\IEEEauthorblockA{
\textit{Department of Electrical and Computer Engineering} \\
\textit{Isfahan University of Technology} \\
Isfahan, Iran \\
}
}

\maketitle

\begin{abstract}
Decoding upper-limb movements from scalp electroencephalography (EEG) is an important problem for brain--computer interfaces (BCIs), but many deep learning models provide limited insight into the brain regions underlying their predictions. We propose a two-branch architecture that combines an LGGNet-inspired local--global graph branch with an EEGNet-style convolutional branch for movement-versus-rest classification. The graph branch groups 61 EEG electrodes into anatomically informed regions and learns their spatial relationships, providing a direct representation of the regions emphasized by the model. We evaluate the proposed architecture using leave-one-subject-out cross-validation on the Ofner upper-limb movement dataset. The model achieves a mean F1-score of $0.74\pm0.06$ across subjects. Beyond classification performance, we analyze the learned graph structure to identify the regions with the strongest connectivity. Across all held-out subjects, the highest connectivity is consistently observed in the \emph{Parietal Left Low} region, corresponding to electrodes P1 and P3. This region is opposite to the performed right-arm movements and lies within the posterior parietal area previously associated with upper-limb movement processing. These results show that incorporating a region-based graph into the decoder can provide an anatomically interpretable representation while maintaining effective movement classification, offering a practical approach toward more transparent EEG-based BCIs.
\end{abstract}


\section{Introduction}

Helping people who have lost motor control is one of the main goals of brain--computer interfaces (BCIs)~\cite{millan2010, muellerputz2019}. Among the available techniques for recording brain activity, EEG is widely used because it is non-invasive, relatively inexpensive, and easy to set up. However, EEG also has some important limitations. Its low spatial resolution and sensitivity to noise make it difficult to extract the information needed to distinguish between different movements.

Ofner \emph{et al.}~\cite{ofner2017} showed that single upper-limb movements can be decoded from slow ($<3$~Hz) EEG signals using movement-related cortical potentials in the time domain, rather than changes in rhythm power. Their analysis also showed that the most useful information mainly came from the premotor cortex, primary motor cortex, primary somatosensory (postcentral) cortex, and posterior parietal cortex.
These findings also provide a useful criterion for evaluating decoding models: in addition to achieving good classification performance, a reliable decoder should primarily capture neural activity from brain regions known to be involved in motor processing.

Deep learning models such as EEGNet~\cite{eegnet} and DeepConvNet~\cite{deepconvnet} have demonstrated strong performance in EEG-based decoding by learning discriminative temporal and spatial filters directly from the input signals. However, the learned filters in conventional convolutional networks are often difficult to interpret in terms of the underlying brain regions. Graph neural networks (GNNs), in contrast, provide a more interpretable representation by modeling EEG electrodes as nodes and their relationships as edges, allowing the learned spatial structure to be examined in relation to brain anatomy. LGGNet~\cite{lggnet}, proposed by Ding \emph{et al.}, introduces local and global graph structures by first grouping electrodes into local brain regions and then modeling interactions between these regions. LGGNet has shown competitive performance across several EEG-based tasks, including attention, fatigue, emotion, and preference recognition. However, its application to upper-limb movement decoding and the relationship between its learned graph structure and motor-related brain anatomy have not been systematically investigated.

Motivated by these complementary strengths, we design our model with two parallel branches: a convolutional branch inspired by EEGNet and a graph-based branch derived from LGGNet~\cite{lggnet}. Convolutional EEG models have generally achieved strong decoding performance, while graph-based models can provide useful spatial representations and greater interpretability. We therefore fuse these two approaches to combine their advantages rather than relying on either representation alone. The convolutional branch is designed to capture discriminative temporal and spatial patterns that contribute to classification performance, whereas the graph branch models relationships between electrodes and brain regions, providing a structure that can be examined from a neurophysiological perspective. We combine the outputs of the two branches to form the final representation. In this way, the proposed architecture aims to achieve effective movement decoding while retaining an interpretable graph-based component that can be evaluated against the known organization of the motor system.
The main contributions of this work are as follows:

\begin{itemize}

\item \textbf{Cross-subject movement decoding.}
We evaluate a dual-branch architecture for upper-limb movement-versus-rest
classification on the Ofner dataset. The model combines convolutional and
graph-based representations, and its cross-subject generalization is assessed
using leave-one-subject-out cross-validation with per-subject F1 scores.

\item \textbf{Interpretable spatial modeling.}
We combine an EEGNet-style convolutional branch with an LGGNet-inspired graph
branch to capture complementary temporal--spatial and inter-electrode
representations. While the convolutional branch supports movement decoding, the
graph branch explicitly models relationships between predefined EEG regions.
We use the learned global adjacency matrix to quantify the connectivity
strength of individual regions.

\item \textbf{Neurophysiological analysis.}
We examine whether the learned spatial representation is consistent with known
neural correlates of upper-limb movement. The strongest connectivity is
consistently observed in the left posterior parietal region, opposite to
the side of the right-arm movements, in agreement with previous EEG findings
~\cite{ofner2017}. This analysis provides a direct neurophysiological
interpretation of the learned graph beyond classification performance.

\end{itemize}

The remainder of this paper is organized as follows. Section~II reviews related
work on EEG-based movement decoding and graph neural networks. Section~III
describes the dataset, proposed architecture, and experimental protocol.
Section~IV presents the classification results, while Section~V analyzes the
learned graph and its neurophysiological interpretation. Section~VI concludes the paper.

\section{Related Work}

Decoding arm movements from scalp EEG has a long history. Early work decoded
imagined wrist and forearm movements from the frequency content of the
signal~\cite{vuckovic2008}. Ofner \emph{et al.}~\cite{ofner2017} then showed
that six executed and imagined upper-limb movements can be decoded from slow
EEG in the time domain. That study provides both the dataset used in this work and the motivation for focusing on motor and sensory areas, where the most useful EEG signals are expected. Later work used
convolutional and recurrent networks on movement-related potentials and
generally found the strongest features just after movement starts, over the
central electrodes above sensorimotor cortex~\cite{zhang2023}. The importance of spatial information in neural signal decoding ~\cite{riseieeg,combinedchannel} also motivates
graph-based representations for scalp EEG.

On the modeling side, compact architectures such as EEGNet~\cite{eegnet} and the deeper DeepConvNet~\cite{deepconvnet} have been widely used for EEG decoding, while TSception~\cite{tsception} introduced multi-scale temporal filters to capture information across different time scales. Although these models have demonstrated strong performance, they have two limitations for our purpose. First, they treat EEG electrodes primarily as a vector or grid and do not explicitly model their anatomical relationships. Second, and more importantly, their learned representations are difficult to interpret in terms of specific brain regions. Their information is distributed across multiple learned filters, requiring additional saliency or visualization methods to infer which brain areas contribute to a decision, and such interpretations may not always be stable. For clinical BCI applications, the lack of a direct link between model predictions and specific brain regions can limit the interpretability and physiological validation of the system. This motivates the use of a model whose learned spatial structure can be directly examined in anatomical terms.

Graph neural networks address the first limitation by representing EEG as a graph, allowing the spatial relationships between electrodes to be explicitly modeled. DGCNN~\cite{dgcnn} and RGNN~\cite{rgnn}, for example, learn connectivity patterns between EEG channels for emotion recognition, with RGNN additionally incorporating neuroscience-inspired priors. LGGNet~\cite{lggnet} extends this approach through a two-level local--global graph structure that explicitly groups electrodes into anatomically defined brain regions. This region-based design makes the learned graph weights easier to interpret in terms of their anatomical meaning. Inspired by this framework, we use the graph component of LGGNet and combine it with an EEGNet-style convolutional branch. Our focus is on applying this combined architecture to movement-versus-rest decoding and, more importantly, on examining whether the learned graph identifies brain regions that are consistent with known motor-related neurophysiology.

\section{Materials and Methods}

\subsection{Dataset}

We use the upper-limb movement dataset introduced by Ofner \emph{et al.}~\cite{ofner2017}.
The dataset contains EEG recordings from 15 healthy participants using 61
electrodes. It includes both motor execution (ME) and motor imagery (MI)
sessions involving six right-arm movements: elbow flexion, elbow extension,
forearm supination, forearm pronation, hand closing, and hand opening, together
with a rest condition.

In this study, we focus on the binary \emph{movement-versus-rest} classification
task, where the goal is to determine whether an EEG trial corresponds to an
upper-limb movement or a resting state, regardless of the specific movement
type. We use only the ME session, in which
participants physically performed the instructed right-arm movements.

Each recording is segmented into fixed, non-overlapping 4~s windows. Each movement trial is represented by a single 4~s window, spanning from 2~s before to 2~s after the movement cue. The remaining rest periods are divided into consecutive, non-overlapping 4~s windows.

\begin{figure*}[t]
\centering
\includegraphics[width=0.98\textwidth]{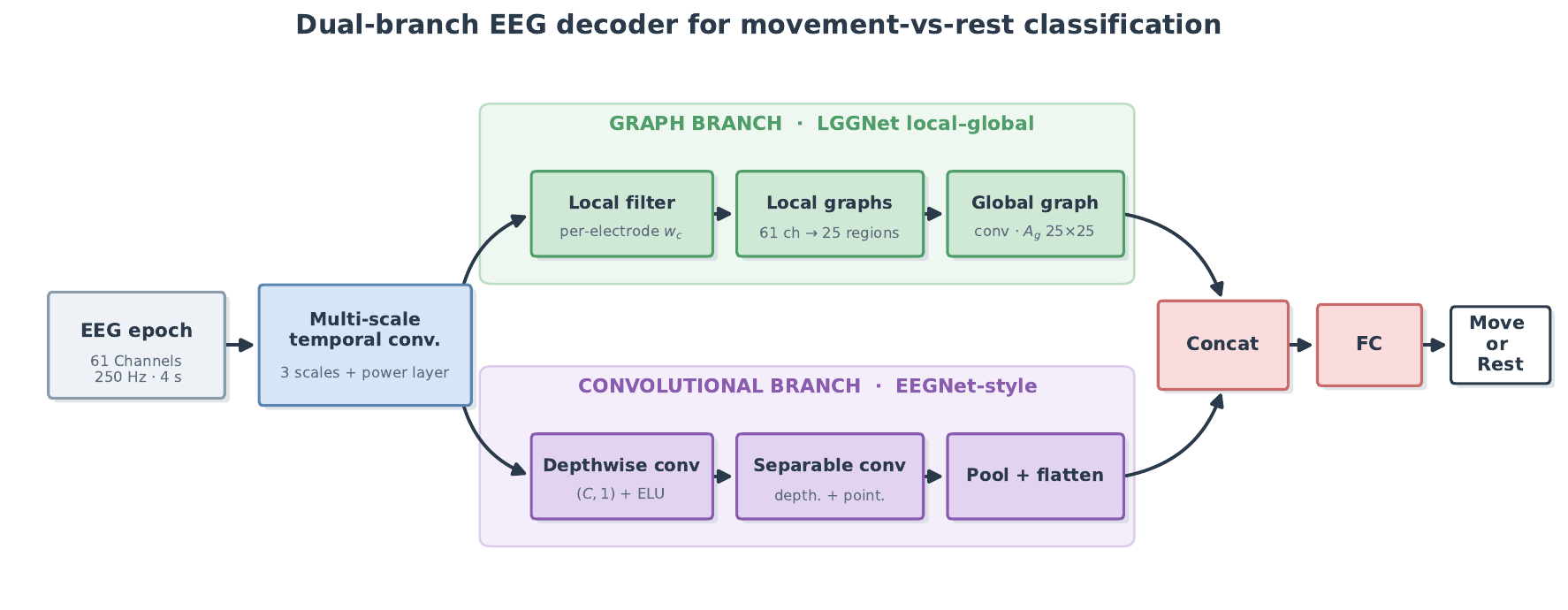}
\caption{Overview of the proposed dual-branch architecture for movement-versus-rest
EEG decoding. A shared temporal feature extraction module feeds an
LGGNet-inspired local--global graph branch and an EEGNet-style convolutional
branch. Their representations are fused and passed to the final classifier.}
\label{fig:arch}
\end{figure*}

\subsection{Model Architecture}
The model takes an EEG trial represented as a tensor
$X\in\mathbb{R}^{1\times C\times L}$, where $C=61$ denotes the number of
electrodes and $L=1000$ denotes the number of temporal samples. With a sampling
frequency of 250~Hz, each trial corresponds to a 4~s window of EEG activity. As illustrated
in Fig.~\ref{fig:arch}, the proposed architecture consists of two parallel
branches: a graph-based branch inspired by the local--global structure of
LGGNet~\cite{lggnet} and a convolutional branch based on the depthwise--separable design of EEGNet~\cite{eegnet}. The two branches process a shared
temporal representation and learn complementary features. The graph branch
models relationships between anatomically grouped EEG regions, while the
convolutional branch extracts discriminative temporal--spatial features. Their
learned representations are subsequently concatenated and provided to a final
classifier.

Because the proposed architecture combines two complementary branches, it
contains more trainable parameters than a single-branch EEG convolutional
network. To reduce the risk of overfitting, particularly given the limited
number of subjects, we use fewer convolutional filters than in the original
EEGNet architecture.

\subsubsection{Shared Temporal Feature Extraction}

Both branches receive features generated by the shared temporal feature
extraction module. To capture temporal patterns at multiple time scales, three
temporal convolutions are applied using kernel lengths of
$\{0.5,0.25,0.125\}\times f_s$ samples, where $f_s=250$~Hz. Each temporal
convolution is followed by a power operation that squares the filtered signal,
applies average pooling over a temporal window, and computes its logarithm:
\begin{equation}
Z_s =
\log\left(\operatorname{AvgPool}\left[(W_s * X)^2\right]\right),
\qquad s\in\{1,2,3\}
\end{equation}
where $W_s$ denotes the temporal convolution kernel at scale $s$. This
operation provides a representation of temporal signal power rather than the
raw EEG waveform.

The outputs of the three temporal scales are concatenated along the temporal
feature dimension and batch-normalized. A $1\times1$ convolution, followed by
a LeakyReLU activation and pooling, is then applied to obtain the shared
feature representation
$U\in\mathbb{R}^{n_T\times C\times d}$, where $n_T$ denotes the number of
temporal filters and $d$ denotes the resulting feature dimension. The resulting
representation is provided to both the graph and convolutional branches.

\subsubsection{Graph Branch (LGGNet-Inspired)}

The graph branch follows the local--global graph formulation of
LGGNet~\cite{lggnet}. First, a learnable local filtering operation is applied
to the features of each electrode. Let $h_c$ denote the feature vector
associated with electrode $c$, obtained from the shared temporal feature
extraction module. The filtered feature representation is defined as
\begin{equation}
\tilde{h}_c =
\operatorname{ReLU}(w_c\odot h_c-b_c)
\end{equation}
where $w_c$ is a learnable element-wise weighting vector, $b_c$ is a learnable
bias vector, and $\odot$ denotes element-wise multiplication. This operation
allows the model to learn the relative importance of the features associated
with each electrode before they are aggregated into spatial regions.

\begin{figure}[t]
\centering
\includegraphics[width=0.92\linewidth]{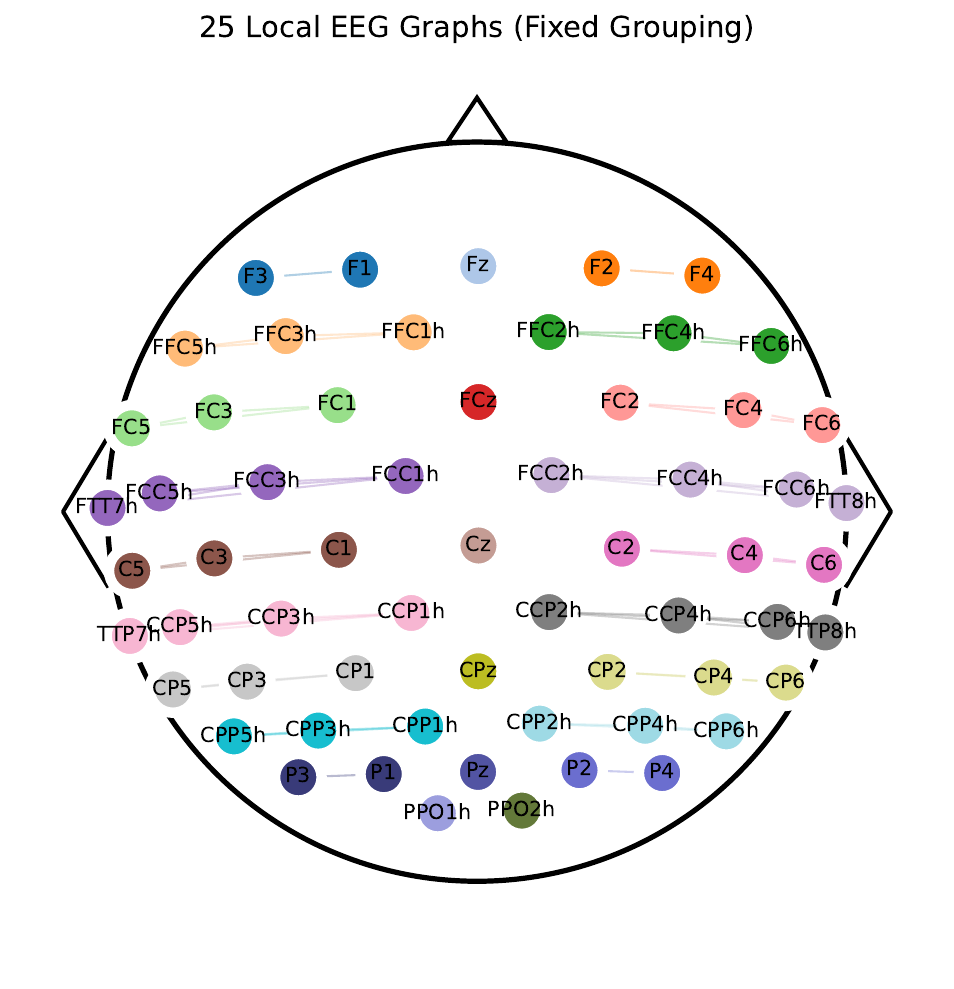}
\caption{Spatial grouping of the 61 EEG electrodes into 25 anatomically informed
regions used by the graph branch. Spatially adjacent electrodes are grouped
according to their scalp locations and channel families. The resulting regions
form the nodes of the global graph.}
\label{fig:electrodes}
\end{figure}

The 61 electrodes are grouped into $R=25$ local regions according to their
spatial positions on the scalp. We use a fixed,
anatomically informed grouping in which spatially adjacent electrodes from the
same channel family are assigned to the same region. For example,
$\{F3,F1\}$ forms the ``Frontal Left'' region,
$\{Fz\}$ forms the ``Frontal Midline'' region,
$\{C5,C3,C1\}$ forms the ``Central Left'' region, and
$\{P1,P3\}$ forms the ``Parietal Left Low'' region. All 25 electrode groups are shown in Fig.~\ref{fig:electrodes}, where each
group is marked with a different color.
The same grouping is
used for all subjects and cross-validation folds and is fixed before model
training, providing a consistent spatial partition of the EEG montage and a
compact set of region nodes for the global graph.

For each region $r$, the features of its electrodes are averaged to obtain a
region-level feature vector:
\begin{equation}
g_r = \frac{1}{N_r}
\sum_{c\in\mathcal{R}_r} \tilde{h}_c,
\qquad r=1,\ldots,R
\end{equation}
where $\mathcal{R}_r$ denotes the set of electrodes assigned to region $r$
and $N_r=|\mathcal{R}_r|$ is the number of electrodes in that region.
The resulting region-level features are then arranged as
\begin{equation}
G=[g_1,\ldots,g_R]\in\mathbb{R}^{R\times d}
\end{equation}
Here, $R=25$ is the number of regions and $d$ is the feature dimension.
Thus, the local graph stage converts the 61 electrode features into 25
spatially defined region representations.

The region representations are then used to construct a global graph that
models interactions between the 25 regions. The graph combines the similarity
between region features with a trainable region-to-region connectivity matrix
$A_g$, resulting in a data-dependent adjacency matrix. This adjacency matrix
is normalized and provided to a graph convolutional network (GCN) layer,
which propagates information between the region nodes. Following the
LGGNet~\cite{lggnet} formulation, the graph convolution can be expressed
compactly as
\begin{equation}
G' =
\operatorname{ReLU}\left(\hat{A}(G\Theta_g-b_g)\right)
\end{equation}
where $\hat{A}$ is the normalized learned adjacency matrix and
$\Theta_g$ and $b_g$ are learnable GCN parameters. The resulting graph
representation is flattened and passed to the feature-fusion stage. The trainable matrix $A_g$ is retained for the interpretability analysis in
Section~\ref{sec:interpret}. Since its rows and columns correspond to the
spatially defined EEG regions, the learned values of $A_g$ provide a direct
representation of the region-to-region relationships emphasized by the graph
branch.

\subsubsection{Convolutional Branch (EEGNet-Inspired)}

In parallel with the graph branch, the shared representation $U$ is processed by
a convolutional branch inspired by the depthwise--separable architecture of
EEGNet~\cite{eegnet}. First, a depthwise convolution is applied across the
electrode dimension to learn spatial filters from the EEG channels. The
resulting features are followed by batch normalization, ELU activation,
average pooling, and dropout. A separable convolution is then applied to
further extract temporal--spatial features while keeping the number of
learnable parameters relatively small. The resulting feature representation is
flattened and passed to the feature-fusion stage. This branch is included to
provide a complementary convolutional representation of the EEG signal and
improve the decoding capability of the combined architecture.

\subsubsection{Feature Fusion and Classification}
The feature vectors produced by the graph and convolutional branches are
concatenated to form a joint representation. Dropout is then applied to the
fused representation, which is subsequently passed through a fully connected
layer to produce two class scores corresponding to movement and rest. The model
is optimized using cross-entropy loss, as described in
the training procedure.

\subsection{Cross-Validation and Training}

We evaluate the model using leave-one-subject-out (LOSO) cross-validation. In
each fold, one subject is held out for testing, while the remaining
14 subjects are used for model development. The development subjects are divided
into training and validation sets using an 80/20 split. The validation set is used for early
stopping and model selection, whereas the held-out test subject is not used
during training or model selection. This protocol evaluates the model's
ability to generalize to unseen subjects.

Training uses the Adam optimizer with a learning rate of $0.001$ and a batch
size of $16$, with early stopping when the validation loss does not improve
for $10$ epochs. All experiments are implemented in PyTorch 2.1.3 and
performed on a Windows system equipped with two NVIDIA GTX 1080 GPUs and
32~GB of RAM.

\section{Results}

\subsection{Overall Performance}

Under the leave-one-subject-out cross-validation protocol, each held-out
subject is evaluated over five independent runs. In each run, the
non-test subjects are randomly divided into training and validation sets using
the specified 80/20 split, resulting in different training and validation
partitions across runs. The F1-score is computed on the held-out subject for each run, and the mean and standard deviation across the five runs are reported for each subject in Table~\ref{tab:persubject}.

The proposed model achieves an average F1 score of $0.74$ across the held-out
subjects. As shown in Table~\ref{tab:persubject}, the mean F1 score varies
across subjects, ranging from $0.61$ for S14 to $0.85$ for S10. The reported standard deviations
reflect the variability caused by the different training and validation
partitions across the five runs. Overall, these results demonstrate the
cross-subject generalization of the proposed architecture. The following
section investigates the learned graph structure to examine which brain
regions contribute most strongly to the model's decisions.

\begin{table}[h]
\caption{Per-subject movement-versus-rest F1 score under leave-one-subject-out
cross-validation. Each value represents the mean $\pm$ standard deviation
across five runs with different training and validation splits.}
\label{tab:persubject}
\centering
\renewcommand{\arraystretch}{1.5}
\setlength{\tabcolsep}{6pt}
\begin{tabular}{lc@{\hspace{2.8em}}lc}
\toprule
Subject & F1 score (Mean$\pm$Std) & Subject & F1 score (Mean$\pm$Std) \\
\midrule
S1 & $0.72\pm0.05$ & S9  & $0.68\pm0.03$ \\
S2 & $0.82\pm0.03$ & S10 & $\mathbf{0.85\pm0.09}$ \\
S3 & $0.76\pm0.08$ & S11 & $0.77\pm0.06$ \\
S4 & $0.77\pm0.04$ & S12 & $0.68\pm0.04$ \\
S5 & $0.74\pm0.06$ & S13 & $0.69\pm0.02$ \\
S6 & $0.81\pm0.02$ & S14 & $0.61\pm0.08$ \\
S7 & $0.66\pm0.07$ & S15 & $0.79\pm0.05$ \\
S8 & $0.69\pm0.05$ & \textbf{Mean} & $\mathbf{0.74\pm0.06}$ \\
\bottomrule
\end{tabular}
\end{table}

\section{Graph Weight Analysis and Physiological Interpretation}
\label{sec:interpret}
The graph branch provides an interpretable representation of the spatial
relationships learned from the EEG signals. We analyze the trained global
adjacency matrix of the graph branch, which describes how strongly the model
links the different brain regions. Each row and column of this matrix
corresponds to one region node, so the learned connectivity can be read in
terms of brain areas rather than single electrodes.

\subsection{From the Learned Graph to a Score per Region}

For each held-out subject, we take the trained global adjacency matrix
$A_g\in\mathbb{R}^{R\times R}$, where $R=25$ denotes the number of regions.
$A_g$ captures model-derived relationships rather than biological
connectivity, and since region features are averaged over electrodes, the
resulting scores do not scale with region size.
Because the model uses the adjacency matrix in symmetric form, we first compute

\begin{equation}
A_s = A_g + A_g^\top
\end{equation}
where $A_s$ is the symmetric adjacency matrix. We then normalize its values to
the range $[0,1]$ for visualization and comparison across subjects:
\begin{equation}
\tilde{A} =
\frac{A_s-\min(A_s)}
{\max(A_s)-\min(A_s)}
\end{equation}

\begin{figure}[t]
\centering
\includegraphics[width=\linewidth]{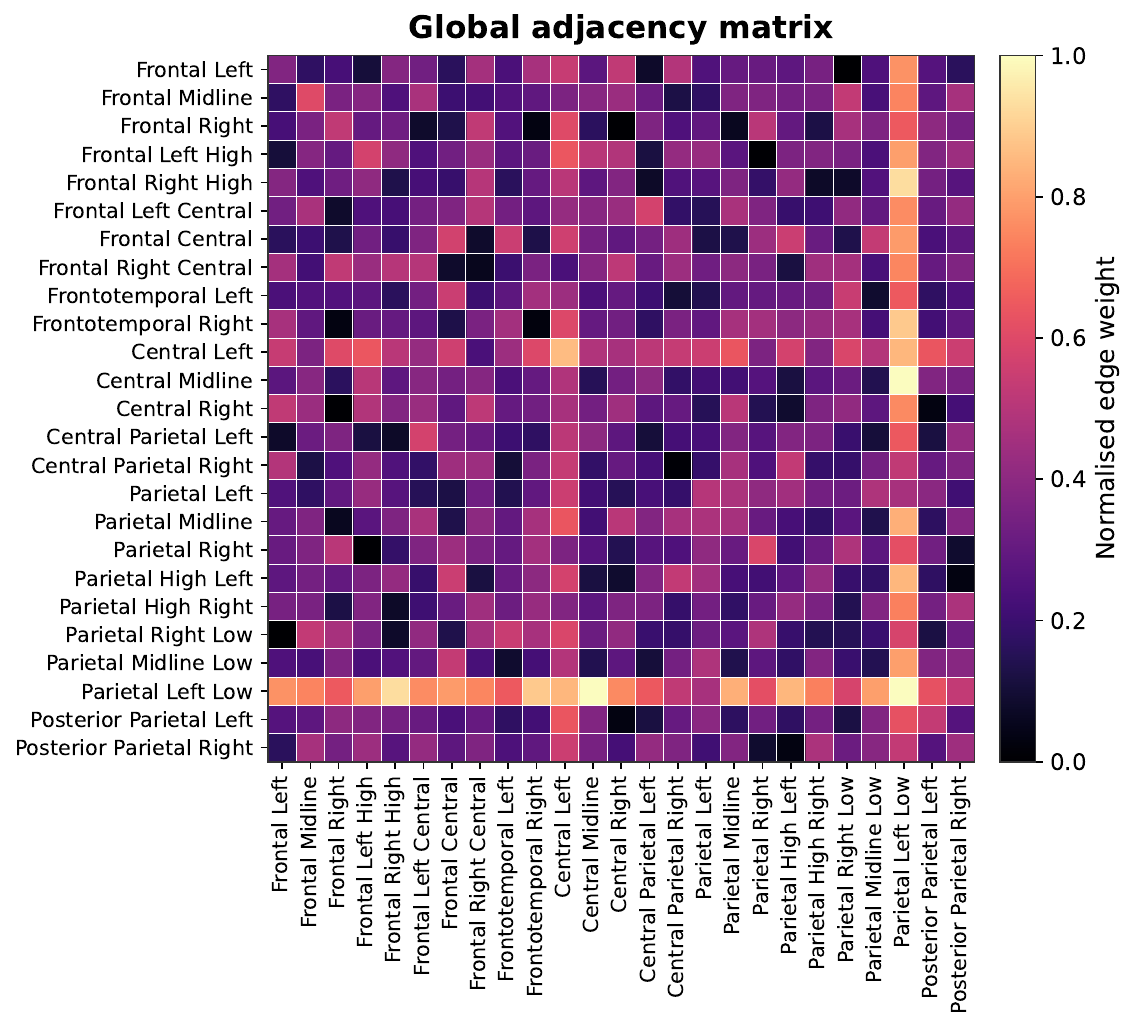}
\caption{Final learned adjacency matrix for a representative held-out subject (S4),
combining the learned spatial relationships from the graph branch. Rows and
columns correspond to the 25 predefined EEG regions, and matrix values
represent the learned connectivity strengths between region pairs.}
\label{fig:adj}
\end{figure}

\begin{figure}[t]
\centering
\includegraphics[width=\linewidth]{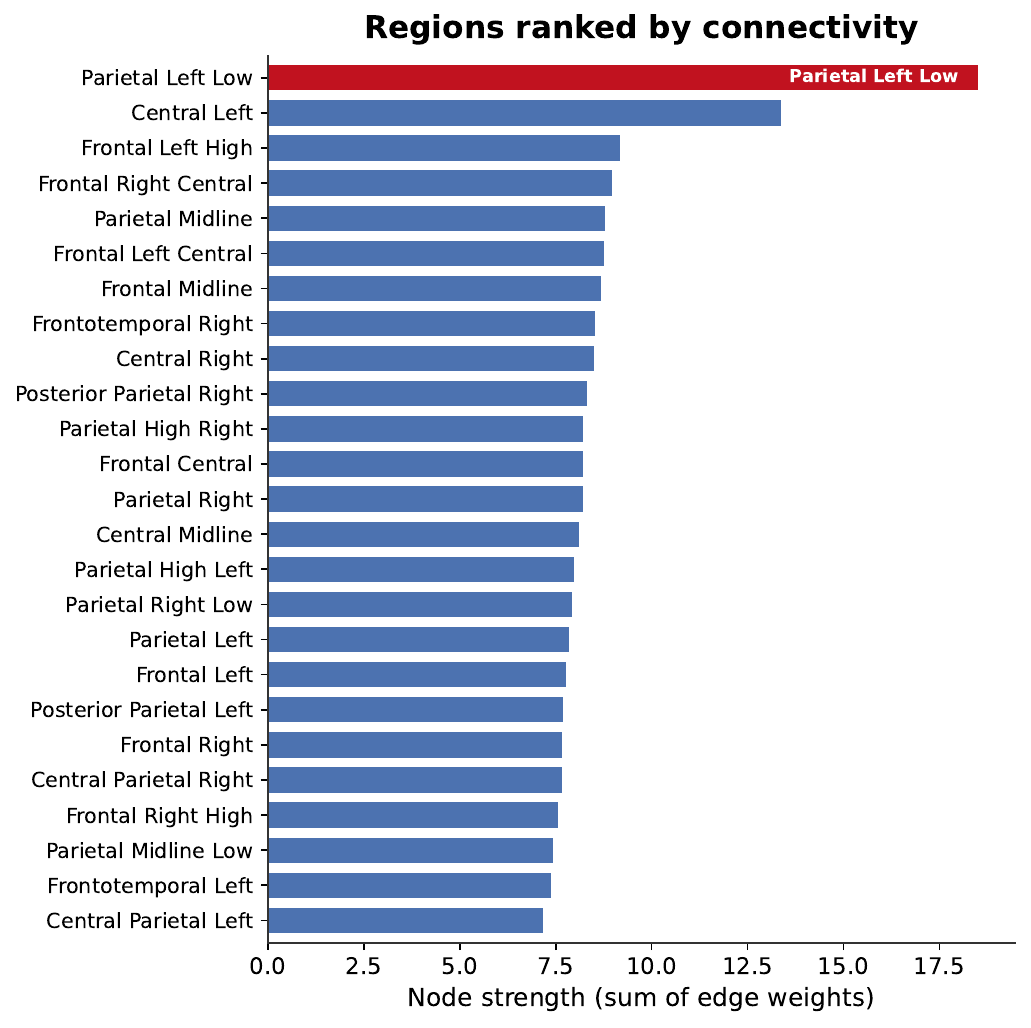}
\caption{Ranking of EEG regions according to their connectivity strength in the
learned graph for a representative held-out subject (S4). Region strength is
computed as the sum of the learned edge weights associated with each region.}
\label{fig:edges}
\end{figure}

The connectivity strength of each region is then calculated as the sum of its
edge weights:
\begin{equation}
s_r = \sum_{j=1}^{R} \tilde{A}_{rj},
\qquad r=1,\ldots,R
\end{equation}
Because $\tilde{A}$ is symmetric, the row and column sums are equivalent.
Thus, $s_r$ represents the overall connectivity strength of region $r$ in the
learned graph. Regions with larger $s_r$ values have stronger connections to
the other regions. Fig.~\ref{fig:adj} shows the learned adjacency matrix $\tilde{A}$ for one
subject, while Fig.~\ref{fig:edges} shows the corresponding ranking of regions
according to their connectivity strength.

\subsection{What the model learned}
The connectivity scores show a clear and consistent pattern across subjects.
The region with the highest connectivity strength is \emph{Parietal Left Low},
formed by the left parietal electrodes P1 and P3 (Fig.~\ref{fig:edges}). This region ranked first for all 15 held-out subjects. One-sided paired
Wilcoxon signed-rank tests comparing its connectivity strength with that of
each of the other 24 regions were all significant (maximum
Bonferroni-adjusted $p=7.3\times10^{-4}$). This indicates that the
left-parietal hub is a consistent property of the task and the trained model
rather than of a particular subject.

\subsection{Match with Brain Physiology}

The learned spatial pattern is consistent with the known neurophysiology of
upper-limb movement. The subjects performed movements with the
\emph{right} arm, which is primarily controlled by the opposite to
\emph{left} cerebral hemisphere. The region emphasized by the model, the
left parietal area around electrodes P1 and P3, is therefore located on the
expected opposite to side. This region also lies within the posterior
parietal area identified by Ofner \emph{et al.}~\cite{ofner2017}, together with
premotor, primary motor, and somatosensory regions, as containing information
related to upper-limb movements. The posterior parietal cortex is involved in
integrating sensory and motor information and contributes to the planning and
guidance of goal-directed movements.

Importantly, the left parietal region was consistently assigned the highest
connectivity strength across the held-out subjects. This consistency suggests
that the learned spatial representation captures a reproducible
movement-related pattern rather than a subject-specific effect. 

The learned pattern is also consistent with the previous invasive
analysis of the same movement-versus-rest task using ECoG data
(RISE-iEEG~\cite{riseieeg}). In that analysis, the postcentral gyrus was among
the regions showing strong discriminative activity. The postcentral gyrus and
posterior parietal cortex are anatomically adjacent and form part of the
broader sensorimotor network involved in movement-related processing.
Therefore, the agreement between the invasive ECoG analysis and the present
scalp-EEG model provides complementary evidence that the learned spatial
representation is related to physiologically meaningful sensorimotor regions.

Overall, the graph-weight analysis identifies a spatial pattern that is
consistent with the expected opposite to organization of right-arm
movement and with previous neurophysiological findings. For BCI applications,
such interpretability can complement classification performance by providing
insight into the brain regions emphasized by the model and helping assess
whether its learned representations are physiologically plausible.

\section{Conclusion}

We developed a two-branch architecture combining an LGGNet-inspired graph
branch with an EEGNet-style convolutional branch for movement-versus-rest
classification on the Ofner upper-limb dataset. Under leave-one-subject-out
cross-validation, the model achieved a mean F1-score of $0.74\pm0.06$. Beyond
classification performance, the learned graph was analyzed to identify the
spatial regions emphasized by the model. Across all held-out subjects, the
highest connectivity strength was consistently observed in the
\emph{Parietal Left Low} region, corresponding to electrodes P1 and P3. This
region is opposite to the performed right-arm movements and lies within
the posterior parietal cortex, which has previously been associated with
upper-limb movement processing~\cite{ofner2017}. The consistency of this
spatial pattern across subjects and its agreement with known neurophysiology
support the interpretability and physiological plausibility of the learned
graph representation. Future work will validate the model's performance and learned spatial patterns across additional datasets.


\begin{thebibliography}{00}

\bibitem{millan2010} J. d. R. Mill\'an \emph{et al.}, ``Combining
brain--computer interfaces and assistive technologies: State-of-the-art and
challenges,'' \emph{Frontiers in Neuroscience}, vol.~4, p.~161, 2010.

\bibitem{muellerputz2019} G. R. M\"uller-Putz \emph{et al.}, ``Towards
noninvasive hybrid brain--computer interfaces: Framework, practice, clinical
application, and beyond,'' \emph{Proc. IEEE}, vol.~103, no.~6,
pp.~926--943, 2015.

\bibitem{riseieeg}
M. O. Memar, N. Ziaei, B. Nazari, and A. Yousefi,
``RISE-iEEG: Robust to inter-subject electrodes implantation variability iEEG classifier,''
in \textit{2025 47th Annual International Conference of the IEEE Engineering in Medicine and Biology Society (EMBC)}, 2025, pp. 1--7.

\bibitem{ofner2017} P. Ofner, A. Schwarz, J. Pereira, and G. R.
M\"uller-Putz, ``Upper limb movements can be decoded from the time-domain of
low-frequency EEG,'' \emph{PLOS ONE}, vol.~12, no.~8, p.~e0182578, 2017.

\bibitem{lggnet} Y. Ding, N. Robinson, C. Tong, Q. Zeng, and C. Guan,
``LGGNet: Learning from local-global-graph representations for
brain--computer interface,'' \emph{IEEE Trans. Neural Netw. Learn. Syst.},
2023. arXiv:2105.02786.

\bibitem{eegnet} V. J. Lawhern, A. J. Solon, N. R. Waytowich, S. M. Gordon,
C. P. Hung, and B. J. Lance, ``EEGNet: A compact convolutional neural network
for EEG-based brain--computer interfaces,'' \emph{J. Neural Eng.}, vol.~15,
no.~5, p.~056013, 2018.

\bibitem{deepconvnet} R. T. Schirrmeister \emph{et al.}, ``Deep learning with
convolutional neural networks for EEG decoding and visualization,'' \emph{Hum.
Brain Mapp.}, vol.~38, no.~11, pp.~5391--5420, 2017.

\bibitem{tsception} Y. Ding, N. Robinson, S. Zhang, Q. Zeng, and C. Guan,
``TSception: Capturing temporal dynamics and spatial asymmetry from EEG for
emotion recognition,'' \emph{IEEE Trans. Affect. Comput.}, vol.~14, no.~3,
pp.~2238--2250, 2023.

\bibitem{dgcnn} T. Song, W. Zheng, P. Song, and Z. Cui, ``EEG emotion
recognition using dynamical graph convolutional neural networks,'' \emph{IEEE
Trans. Affect. Comput.}, vol.~11, no.~3, pp.~532--541, 2020.

\bibitem{rgnn} P. Zhong, D. Wang, and C. Miao, ``EEG-based emotion recognition
using regularized graph neural networks,'' \emph{IEEE Trans. Affect. Comput.},
vol.~13, no.~3, pp.~1290--1301, 2022.

\bibitem{vuckovic2008} A. Vu\v{c}kovi\'c and F. Sepulveda, ``Delta band
contribution in cue based single trial classification of real and imaginary
wrist movements,'' \emph{Med. Biol. Eng. Comput.}, vol.~46, no.~6,
pp.~529--539, 2008.

\bibitem{combinedchannel}
M. O. Memar, N. Ziaei, and B. Nazari, ``A Combined Channel Approach for Decoding Intracranial EEG Signals: Enhancing Accuracy Through Spatial Information Integration,'' in \textit{2025 33rd International Conference on Electrical Engineering (ICEE)}, pp. 363--368, 2025.

\bibitem{zhang2023}
R. Zhang \emph{et al.}, ``Recognition of single upper limb motor imagery tasks
from EEG using multi-branch fusion convolutional neural network,''
\emph{Frontiers in Neuroscience}, vol.~17, p.~1129049, 2023.

\end{thebibliography}
\end{document}